\documentclass{article}

\usepackage{microtype}
\usepackage{graphicx}
\usepackage{subcaption}
\usepackage{booktabs} 

\usepackage{hyperref}

\usepackage[accepted]{icml2026}

\usepackage{amsmath}
\usepackage{amssymb}
\usepackage{mathtools}
\usepackage{amsthm}

\usepackage[capitalize,noabbrev]{cleveref}

\usepackage{tcolorbox}
\usepackage{booktabs}
\usepackage{multirow}
\usepackage{colortbl} 
\usepackage[table]{xcolor}
\usepackage{enumitem}
\usepackage{xcolor}   
\usepackage{wrapfig}
\usepackage{shadowtext}

\usepackage{xcolor}
\usepackage{amsmath}
\usepackage{tcolorbox}
\usepackage{wrapfig}

\definecolor{stdblue}{RGB}{30, 100, 180}
\definecolor{cotgreen}{RGB}{30, 140, 90}
\definecolor{hcotorange}{RGB}{200, 90, 20}
\definecolor{instblue}{RGB}{20, 80, 160}
\definecolor{execpurple}{RGB}{120, 30, 140}
\definecolor{boxbg}{RGB}{248, 249, 252}

\newtcolorbox{stdbox}{%
  colframe=stdblue, colback=boxbg, coltitle=white,
  colbacktitle=stdblue, fonttitle=\bfseries\small,
  title={Standard Prompting},
  boxrule=0.6pt, arc=3pt, titlerule=0pt,
  left=4pt, right=4pt, top=3pt, bottom=3pt
}
\newtcolorbox{cotbox}{%
  colframe=stdblue, colback=boxbg, coltitle=white,
  colbacktitle=stdblue, fonttitle=\bfseries\small,
  title={Chain-of-Thought (CoT) Prompting},
  boxrule=0.6pt, arc=3pt, titlerule=0pt,
  left=4pt, right=4pt, top=3pt, bottom=3pt
}
\newtcolorbox{psbox}{%
  colframe=stdblue, colback=boxbg, coltitle=white,
  colbacktitle=stdblue, fonttitle=\bfseries\small,
  title={Plan-and-Solve Prompting},
  boxrule=0.6pt, arc=3pt, titlerule=0pt,
  left=4pt, right=4pt, top=3pt, bottom=3pt
}
\newtcolorbox{hcotvbox}{%
  colframe=stdblue, colback=boxbg, coltitle=white,
  colbacktitle=stdblue, fonttitle=\bfseries\small,
  title={Hi-CoT (format-relaxed) Prompting},
  boxrule=0.6pt, arc=3pt, titlerule=0pt,
  left=4pt, right=4pt, top=3pt, bottom=3pt
}
\newtcolorbox{hcotbox}{%
  colframe=hcotorange, colback=boxbg, coltitle=white,
  colbacktitle=hcotorange, fonttitle=\bfseries\small,
  title={Hi-CoT Prompting},
  boxrule=0.6pt, arc=3pt, titlerule=0pt,
  left=4pt, right=4pt, top=3pt, bottom=3pt
}

\newcommand{\instag}{\textcolor{instblue}{\texttt{<|instruction|>}}\;}
\newcommand{\extag}{\textcolor{execpurple}{\texttt{<|execution|>}}\;}

\theoremstyle{plain}

\theoremstyle{definition}

\theoremstyle{remark}

\usepackage[textsize=tiny]{todonotes}

\icmltitlerunning{Hi-CoT: Enhancing LLM Reasoning Performance and Efficiency}

\begin{document}

\twocolumn[
  \icmltitle{Hierarchical Chain-of-Thought: \\Enhancing LLM Reasoning Performance and Efficiency}



  \icmlsetsymbol{equal}{*}

  \begin{icmlauthorlist}
    \icmlauthor{Xingshuai Huang}{comp}
    \icmlauthor{Derek Li}{comp1}
    \icmlauthor{Bahareh Nikpour}{comp}
    \icmlauthor{Parsa Omidi}{comp}
  \end{icmlauthorlist}

  \icmlaffiliation{comp}{Huawei Technologies Canada, Canada}
  \icmlaffiliation{comp1}{Huawei Noah’s Ark Lab, Canada}

  \icmlcorrespondingauthor{Xingshuai Huang}{xingshuai.huang@gmail.com}

  \icmlkeywords{Machine Learning, ICML}

  \vskip 0.3in
]



\printAffiliationsAndNotice{}  

\begin{abstract}
Chain-of-Thought (CoT) prompting has significantly improved the reasoning capabilities of large language models (LLMs). However, conventional CoT often relies on unstructured, flat reasoning chains that suffer from redundancy and suboptimal performance.
In this work, we introduce Hierarchical Chain-of-Thought (\textbf{Hi-CoT}), a structured reasoning paradigm specifically designed to address the challenges of complex, multi-step reasoning. Hi-CoT decomposes the reasoning process into hierarchical substeps by alternating between instructional planning and step-by-step execution. 
This decomposition enables LLMs to better manage long reasoning horizons and maintain logical coherence.
Extensive evaluations across diverse LLMs and mathematical reasoning benchmarks show that Hi-CoT consistently improves average accuracy by 6.2\% (up to 61.4\% on certain models and tasks) while reducing reasoning trace length by 13.9\% compared to CoT. We further show that accuracy and efficiency are maximized when models strictly adhere to the hierarchical structure.
Our code is available at https://github.com/XingshuaiHuang/Hi-CoT.
\end{abstract}

\section{Introduction}
\label{sec:intro}

\begin{figure}
  \centering
  \includegraphics[width=\linewidth]{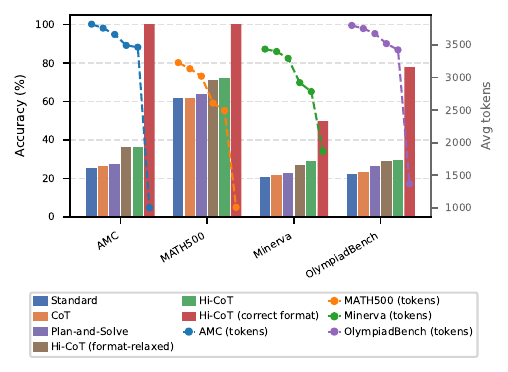}
  \caption{Accuracy (\%) and average token length for Qwen3-8B model \citep{yang2025qwen3} across multiple prompting methods on mathematical reasoning benchmarks. For Hi-CoT (correct format), results are computed only from responses that strictly follow the hierarchical structure.}
  \label{fig:qwen_results}
\end{figure}

Large Language Models (LLMs) have shown strong capabilities on reasoning-intensive tasks when paired with appropriate prompting strategies. One influential approach is Chain-of-Thought (CoT) prompting \citep{wei2022chain}, which encourages models to produce intermediate reasoning steps before a final answer. This simple modification has led to notable gains on multi-step reasoning tasks\citep{puri2021codenet, zhou2025agentfly}, such as mathematic tasks \citep{cobbe2021training, liu2025metascale}.

Despite its success, standard CoT prompting has a strcutural shortcoming. The generated reasoning traces are typically linear and unstructured, which can introduce redundant steps (e.g., repeating explanations) and occasional deviations from the intended line of reasoning \citep{wang2023plan}. Redundancy arises because CoT imposes no compression presssure on the reasoning process. Without an explicit mechanism to filter low-value content, the model can repeat itself, hedge, or wander without penalty. Longer traces do not imply better reasoning \citep{liu2025understanding}; they often reflect disorganized exploration rather than deliberate problem solving, while simultaneously incurring higher inference costs as computational overhead of transformer models grows with the number of generated tokens\citep{zeng2026lethe, omidi2025memory}.  This lack of structure is particularly costly for problems that require coordinated planning across multiple steps, where unguided reasoning frequently fails to maintain logical coherence\citep{he2024olympiadbench, lewkowycz2022solving}.

One solution is to impose a plan. Plan-and-Solve prompting \citep{wang2023plan} does exactly this, separating reasoning into a global planning phase followed by step-by-step solution. While this improves flat CoT, it suffers from a fundamental weakness: the plan is fixed upfront and there is no control over individual execution steps. As the reasoning unfolds, the model may drift from its plan, skip steps, or execute inconsistently (a phenomenon we call plan–execution drift). This happens because the model receives no signal to pause, reassess, or compress its current state before each step. Human experts, by contrast, neither reason in flat chains nor commit to a single rigid plan. Instead, they continuously refine their approach at each step (what should I do next, given where I am?) and local execution (carry it out precisely). Each planning step forces compression of the current reasoning state into a targeted subgoal, filtering out accumulated noise before proceeding \citep{hu2025divide}. 

Motivated by this insight, \textbf{we propose Hierarchical Chain-of-Thought (Hi-CoT) prompting, a structured reasoning paradigm that organizes the reasoning process as a sequence of alternating \textit{instruction} and \textit{execution} steps}, to improve LLM performance and efficiency on complex reasoning tasks. In contrast to Plan-and-Solve, which adopts a fully separated plan–solution structure, Hi-CoT enables fine-grained and adaptive planning/instruction. Each instruction step is conditioned on the outcome of the previous execution, enabling continuous plan refinement rather than sticking blindly to an upfront plan. Hi-CoT operates entirely at inference time and requires no changes to model parameters or architecture. This structure serves as a sequence of compression bottlenecks: before each execution step, the model must distill the current reasoning state into a concise, targeted instruction. This bottleneck mechanism has two effects. First, it filters low-information content from the reasoning trajectory, reducing redundancy and token waste and yeilding a more efficient reasoning process. Second, it keeps each execution step explicitly grounded in a stated goal, preventing drift and maintaining logical coherence across long reasoning horizons.

We evaluate Hi-CoT extensively across 13 model configurations spanning the Qwen3 and DeepSeek-R1 families, and five mathematical reasoning benchmarks of varying difficulty. Our results demonstrate that Hi-CoT provides a rare combination of improvements: better accuracy and lower inference cost simultaneously. It is a non-trivial outcome, as many prior methods for improving reasoning quality increase inference cost by generating longer or multiple reasoning paths (e.g., Tree-of-Thoughts \citep{yao2023tree}, Self-Consistency \citep{wang2022self}.
Our main contributions are:
\begin{itemize} [itemsep=1pt, topsep=2pt]
   \item We propose Hi-CoT, a zero-shot inference-time framework that enforces hierarchical reasoning via compression bottlenecks, without fine-tuning, extra models, or search.
   \item We show that Hi-CoT consistently improves average accuracy by 6.2\% (up to 61.4\%) while reducing reasoning trace length by 13.9\% across 13 models and 5 benchmarks, showing that structure can also be a primary bottleneck in LLM reasoning.
\item We demonstrate that when models strictly stick to the hierarchical format, accuracy reaches 100\% on AMC and MATH500 with traces shortened by up to 75\%, suggesting that current LLMs substantially underutilize their latent reasoning capacity under unstructured prompting. 
\end{itemize}
\section{Related Work}
\label{sec:related_work}

\subsection{Chain-of-Thought Prompting}
Chain-of-Thought (CoT) prompting \citep{wei2022chain} improves LLM reasoning by encouraging intermediate steps before a final answer. Zero-Shot CoT \citep{kojima2022large} shows this can be elicited without annotated examples, while Self-Consistency \citep{wang2022self} improves robustness by sampling multiple reasoning paths. Auto-CoT \citep{zhang2022automatic} automates the construction of few-shot demonstrations through sampling and pseudo-labeling, and Auto-CCoT \citep{li2025automatic} further strengthens this by selecting valid–invalid reasoning pairs to reinforce correct reasoning. \citet{cheng2025revisiting} shows that for modern LLMs, CoT exemplars primarily serve to enforce output format rather than improve reasoning quality. 

More recently, Auto-CCoT \citep{li2025automatic} improves LLM reasoning by automatically generating and selecting informative valid–invalid reasoning pairs from model outputs to reinforce correct reasoning and avoid common errors.
Among efficiency-focused extensions, Chain of Draft (CoD) \citep{xu2025chain} produces concise, information-dense reasoning traces to reduce latency, but this efficiency comes at the expense of model performance.
More recent work has explored automatic prompt optimization, including GAN-CoT \citep{wang2026chain}, which iteratively refines CoT templates via generative adversarial training, and Select-Prompt \citep{che2026select}, which improves reasoning by mining hard samples and optimizing prompts through selecting correct reasoning chains from multiple generated candidates. 
Despite these advances, the lack of structure at the step level leaves models vulnerable to redundant and poorly organized reasoning.


%

\subsection{Structured LLM Reasoning}
Recent work has explored ways of introducing structure into LLM reasoning. One line of research focuses on decomposing complex problems into smaller, more manageable components. Least-to-Most prompting \cite{zhou2022least} breaks a problem into a sequence of simpler subproblems, which are solved incrementally using previously derived results. However, this method requires two stages of manual prompting and mandates task-specific exemplar design for both stages.
Plan-and-Solve Prompting \citep{wang2023plan} improves zero-shot CoT by first planning task decomposition and then solving subtasks step-by-step. However, this method relies on a single upfront plan with weak step-level constraints, which can lead to plan–execution drift, missing steps, and inconsistent reasoning.


Another direction extends the linear reasoning process of CoT to more flexible structures. Tree-of-Thoughts (ToT) \cite{yao2023tree} and Graph-of-Thoughts (GoT) \cite{besta2024graph} organize reasoning as trees or directed graphs, allowing for branching, backtracking, and aggregation of intermediate results.
Similarly, Layer-of-Thoughts (LoT) \citep{fungwacharakorn2024layer} introduces a layered reasoning process in which candidate solutions are progressively filtered and refined. 
Multi-Path Plan Aggregation (MPPA) \citep{xiong2025enhancing} improves long CoT reasoning by generating and aggregating multiple candidate plans during inference.

From a reinforcement learning perspective \citep{huang2024context}, REMA \citep{wan2025rema} and GLIDER \citep{hu2025divide} use hierarchical multi-agent systems with a high-level planner and low-level executor trained jointly to improve performance. 
In-context Decision Transformer (IDT) \citep{huang2024context} adopts a hierarchical in-context RL framework that generates high-level decisions to guide low-level actions and learns to infer these decisions from behavior.
ReasonFlux \citep{yang2025reasonflux} builds on hiearchical RL and extracts reusable reasoning templates. 

However, all these approaches introduce significant overhead through search procedures, multiple inference passes, or additional agents, and offer limited interpretability over individual reasoning steps.
Hi-CoT takes a different path where a single reasoning trajectory with no additional agents or search, enforcing hierarchical organization through a simple alternating instruction–execution structure. 
Unlike Plan-and-Solve \citep{wang2023plan}, it applies step-level constraints continuously, preventing drift and yielding interpretable, auditable reasoning traces.
\section{Methodology}
\label{sec:method}

This section presents the design and implementation of Hierarchical Chain-of-Thought (Hi-CoT) prompting. 
The central goal of Hi-CoT is to introduce structure, goal orientation, and efficiency into the reasoning process, addressing key limitations of CoT prompting.

\begin{figure*}[ht]
  \centering
  \includegraphics[width=\textwidth]{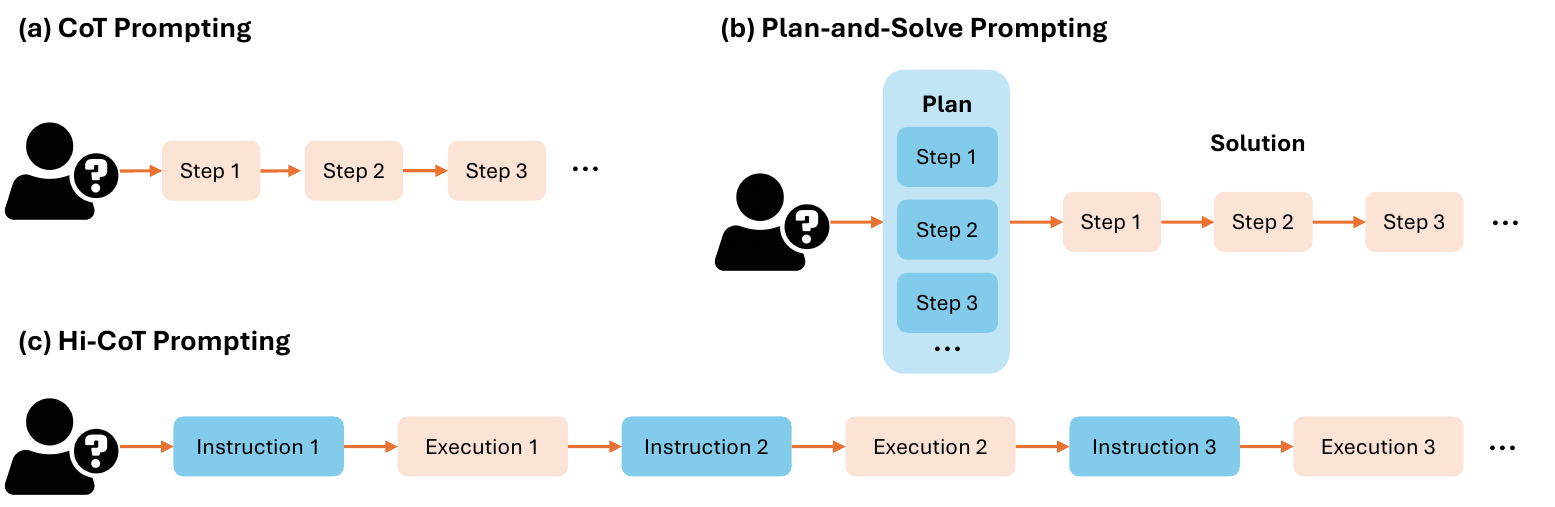}
  \caption{Comparison of reasoning methods. 
  (a) CoT generates step-by-step reasoning, (b) Plan-and-Solve first creates a global plan and then executes it at a coarse level, while (c) Hi-CoT enforces fine-grained alternating instruction–execution steps for more structured and controlled reasoning.
  }
  \label{fig:hicot}
\end{figure*}

Hi-CoT is grounded in the principle of hierarchical reasoning, drawing inspiration from hierarchical RL \citep{yang2025reasonflux, wan2025rema, hu2025divide}, where decision-making is organized across multiple levels of abstraction. In such frameworks, high-level policies define subgoals or plans, while low-level policies execute actions to achieve them. Hi-CoT adopts a similar perspective by separating reasoning into two complementary stages: high-level planning (instruction steps) and low-level execution (execution steps). This design is consistent with recent findings that LLMs exhibit stronger reasoning when explicitly guided through structured subgoals \citep{yang2025reasonflux, wan2025rema, wang2025emergent}, as hierarchical organization constrains the search space and reduces unnecessary exploration.

Specifically, Hi-CoT organizes the reasoning process as a sequence of alternating \texttt{<|instruction|>} and \texttt{<|execution|>} blocks. Each instruction step specifies the immediate objective or plan, and each execution step carries out that plan. The prompt explicitly defines this format, encouraging the model to reason in a structured and deliberate manner rather than producing a single, unstructured chain of thoughts. 
To better situate our approach, we contrast Hi-CoT with the standard CoT prompting and a structured CoT method, namely Plan-and-Solve, as shown in Figure \ref{fig:hicot}.

\textbf{CoT prompting} \citep{wei2022chain, kojima2022large} ($(a)$ in Figure \ref{fig:hicot}) extends standard prompting by explicitly encouraging the model to produce intermediate reasoning steps before arriving at a final answer. 
By externalizing the reasoning process, CoT enables the model to decompose complex problems into a sequence of simpler steps, leading to improved performance on tasks that require multi-step reasoning.
However, it lacks explicit structural constraints to distinguish essential reasoning from filler, often leading to redundant and unfocused outputs.

\textbf{Plan-and-Solve prompting} \citep{wang2023plan} ($(b)$ in Figure \ref{fig:hicot}) introduces a coarse-grained structure by separating reasoning into two stages: planning and solution. While this approach enforces structured reasoning, it relies on a single upfront plan and provides weak step-level constraints, which can lead to plan–execution drift, omitted steps, and inconsistent reasoning during execution.

\textbf{Hi-CoT prompting} ($(c)$ in Figure \ref{fig:hicot}) decomposes a complex reasoning task $T$ into a sequence of $n$ reasoning stages $S_1, S_2, \ldots, S_n$, where $n$ is not fixed in advance but determined dynamically by the model based on the complexity of the problem. Each stage $S_i$ consists of two interdependent components:

\begin{itemize}
\item \textbf{Instruction Step} $I_i$: a high-level plan that specifies the objective and strategy for the next reasoning action. It reflects on the current reasoning state, identifies the gap to the task goal, and proposes a targeted approach to bridge it, enabling adaptive plan refinement rather than relying on a single upfront decomposition.

\item \textbf{Execution Step} $E_i$: a concrete operation that implements the plan in $I_i$, performing the necessary computations or logical inferences. Each execution is explicitly guided by its corresponding instruction, ensuring goal-directed reasoning.
\end{itemize}

  \vspace{3pt}
\begin{hcotbox}
\small
\textbf{Prompt:} Please reason hierarchically by alternating between \texttt{<|instruction|>} (planning what to do next) and \texttt{<|execution|>} (executing that plan).

Each \texttt{<|instruction|>} describes the reasoning step or plan, and each \texttt{<|execution|>} performs the corresponding reasoning or computation.    

You should start with an instruction and follow this format strictly:    

\texttt{<|instruction|>} Step 1: ...    

\texttt{<|execution|>} Step 1: ...    

\texttt{<|instruction|>} Step 2: ...    

\texttt{<|execution|>} Step 2: ...    

...    

Finally, put your final answer within \texttt{\textbackslash boxed$\left\{\right\}$}.\\[2pt]
\textit{A store sells apples for \$3 each and oranges for \$5 each.
If Alice buys 4 apples and 2 oranges, how much does she pay in total?}
 
\tcblower
\small
\textbf{Output:}\\[3pt]
\instag Step 1: Compute the total cost of the apples purchased.\\
\extag  Step 1: $4 \times \$3 = \$12$.
 
\smallskip
\instag Step 2: Compute the total cost of the oranges purchased.\\
\extag  Step 2: $2 \times \$5 = \$10$.
 
\smallskip
\instag Step 3: Sum the two subtotals to obtain the overall cost.\\
\extag  Step 3: $\$12 + \$10 = \$22$.
 
\smallskip
\instag Step 4: State the final answer.\\
\extag  Step 4: The total amount Alice pays is \(\boxed{\$22}\).
\end{hcotbox}

The resulting sequence $(I_1 \rightarrow E_1 \rightarrow I_2 \rightarrow E_2 \rightarrow \cdots \rightarrow I_n \rightarrow E_n)$ forms a hierarchical reasoning trajectory. Each instruction step $I_i$ is conditioned on the outcome of the previous execution step $E_{i-1}$, while each execution step $E_i$ directly follows the clear plan specified in $I_i$. 
Unlike Plan-and-Solve \citep{wang2023plan}, which depends on a single global plan and offers limited control during execution, Hi-CoT introduces fine-grained, step-level planning that continuously updates throughout the reasoning process. This mitigates plan–execution drift and reduces the risk of missing or inconsistent steps.

Moreover, the alternating structure introduces an explicit bottleneck at each stage: the model must compress the current reasoning state into a concise instruction before proceeding. This filters out low-information content and leads to more focused and efficient reasoning. It also improves interpretability, as each action is explicitly grounded in a preceding plan, making the overall reasoning process easier to follow and analyze.
Overall, Hi-CoT establishes a feedback loop that is both adaptive and structured, ensuring consistent alignment between planning and execution.
We show a Hi-CoT prompting example above.


\section{Experiments}
\label{sec:exp}

\definecolor{rowgray}{RGB}{230,230,230}
\begin{table*}
\renewcommand{\arraystretch}{0.95}
\footnotesize

  \centering
  \caption{Accuracy (\%) across 13 models and three prompting strategies. Hi-CoT consistently outperforms baseline prompting methods across all five mathematical reasoning benchmarks.}
  \label{tab:model_eval_results}
  \resizebox{\textwidth}{!}{%
  \begin{tabular}{ll|rrrrr|r}
    \toprule
    \textbf{Model} & \textbf{Prompting} & \textbf{AIME24} & \textbf{AMC} & \textbf{MATH500} & \textbf{Minerva} & \textbf{OlympiadBench} & \textbf{Avg.} \\
    \midrule

    \multirow{5}{*}{\textbf{\texttt{Qwen3-0.6B}}} & Standard & 0.0 & 21.7 & 56.2 & 18.0 & 20.7 & 23.3 \\
     & CoT & 3.3 & 27.7 & 55.4 & 15.8 & 20.7 & 24.6 \\
     & Plan-and-Solve & 3.3 & 26.5 & 56.4 & 16.9 & 20.7 & 24.8 \\
     & Hi-CoT (format-relaxed) & \textbf{6.7} & 25.3 & \textbf{60.0} & 18.0 & 23.0 & 26.6 \\
    \rowcolor{rowgray}
     & Hi-CoT & 3.3 & \textbf{33.7} & 57.0 & \textbf{20.2} & \textbf{23.4} & \textbf{27.5} \\
    \midrule
    \multirow{5}{*}{\textbf{\texttt{Qwen3-1.7B}}} & Standard & 0.0 & 25.3 & 58.2 & 23.9 & 21.2 & 25.7 \\
     & CoT & 3.3 & 32.5 & 63.0 & 21.3 & 22.9 & 28.6 \\
     & Plan-and-Solve & 3.3 & 27.7 & 63.2 & 21.7 & 25.8 & 28.3 \\
     & Hi-CoT (format-relaxed) & 6.7 & 36.1 & 73.0 & \textbf{30.1} & 32.0 & 35.6 \\
    \rowcolor{rowgray}
     & Hi-CoT & \textbf{10.0} & \textbf{38.6} & \textbf{75.2} & 27.6 & \textbf{32.3} & \textbf{36.7} \\
    \midrule
    \multirow{5}{*}{\textbf{\texttt{Qwen3-4B-Instruct-2507}}} & Standard & 30.0 & 62.7 & 86.2 & 31.2 & 53.8 & 52.8 \\
     & CoT & 26.7 & 65.1 & 88.8 & 35.3 & 55.1 & 54.2 \\
     & Plan-and-Solve & 26.7 & 65.1 & 87.4 & 34.9 & 53.9 & 53.6 \\
     & Hi-CoT (format-relaxed) & 23.3 & 60.2 & \textbf{90.0} & 34.2 & \textbf{56.3} & 52.8 \\
    \rowcolor{rowgray}
     & Hi-CoT & \textbf{36.7} & \textbf{66.3} & 88.8 & \textbf{36.0} & 56.0 & \textbf{56.8} \\
    \midrule
    \multirow{5}{*}{\textbf{\texttt{Qwen3-4B-Thinking-2507}}} & Standard & 0.0 & 13.3 & 47.4 & 26.1 & 11.4 & 19.6 \\
     & CoT & 0.0 & 12.0 & 46.4 & 30.1 & 10.1 & 19.7 \\
     & Plan-and-Solve & 0.0 & 16.9 & 53.8 & 31.6 & 15.7 & 23.6 \\
     & Hi-CoT (format-relaxed) & \textbf{3.3} & 21.7 & 64.8 & \textbf{35.7} & 26.4 & 30.4 \\
    \rowcolor{rowgray}
     & Hi-CoT & \textbf{3.3} & \textbf{25.3} & \textbf{67.6} & 35.3 & \textbf{27.7} & \textbf{31.8} \\
    \midrule
    \multirow{5}{*}{\textbf{\texttt{Qwen3-4B}}} & Standard & 3.3 & 27.7 & 62.8 & 22.8 & 23.1 & 27.9 \\
     & CoT & 6.7 & 31.3 & 67.6 & 27.2 & 27.6 & 32.1 \\
     & Plan-and-Solve & 10.0 & 30.1 & 66.6 & 25.7 & 27.3 & 31.9 \\
     & Hi-CoT (format-relaxed) & 3.3 & \textbf{42.2} & 74.6 & \textbf{29.8} & 33.9 & 36.8 \\
    \rowcolor{rowgray}
     & Hi-CoT & \textbf{16.7} & 39.8 & \textbf{75.2} & \textbf{29.8} & \textbf{35.4} & \textbf{39.4} \\
    \midrule
    \multirow{5}{*}{\textbf{\texttt{Qwen3-8B}}} & Standard & 0.0 & 25.3 & 62.0 & 21.0 & 22.1 & 26.1 \\
     & CoT & 6.7 & 26.5 & 61.8 & 21.7 & 23.4 & 28.0 \\
     & Plan-and-Solve & 6.7 & 27.7 & 63.8 & 22.8 & 26.5 & 29.5 \\
     & Hi-CoT (format-relaxed) & 10.0 & \textbf{36.1} & 71.4 & 27.2 & 29.2 & 34.8 \\
    \rowcolor{rowgray}
     & Hi-CoT & \textbf{13.3} & \textbf{36.1} & \textbf{72.2} & \textbf{29.0} & \textbf{29.6} & \textbf{36.1} \\
    \midrule
    \multirow{5}{*}{\textbf{\texttt{Qwen3-14B}}} & Standard & 6.7 & 28.9 & 66.0 & 27.6 & 25.6 & 31.0 \\
     & CoT & 3.3 & 27.7 & 69.2 & 26.1 & 27.7 & 30.8 \\
     & Plan-and-Solve & 10.0 & 27.7 & 69.8 & 27.9 & 28.3 & 32.7 \\
     & Hi-CoT (format-relaxed) & 16.7 & \textbf{42.2} & 76.2 & 33.1 & 32.0 & 40.0 \\
    \rowcolor{rowgray}
     & Hi-CoT & \textbf{23.3} & \textbf{42.2} & \textbf{76.6} & \textbf{33.8} & \textbf{33.5} & \textbf{41.9} \\
    \midrule
    \multirow{5}{*}{\textbf{\texttt{Qwen3-32B}}} & Standard & 10.0 & 30.1 & 66.4 & 27.9 & 29.8 & 32.8 \\
     & CoT & \textbf{13.3} & 36.1 & 70.0 & 32.4 & 29.5 & 36.3 \\
     & Plan-and-Solve & \textbf{13.3} & 36.1 & 70.4 & 29.0 & 32.9 & 36.4 \\
     & Hi-CoT (format-relaxed) & \textbf{13.3} & 36.1 & \textbf{76.8} & 35.7 & 34.8 & 39.4 \\
    \rowcolor{rowgray}
     & Hi-CoT & 10.0 & \textbf{44.6} & \textbf{76.8} & \textbf{37.1} & \textbf{35.0} & \textbf{40.7} \\
    \midrule
    \multirow{5}{*}{\textbf{\texttt{DeepSeek-R1-Distill-Qwen-1.5B}}} & Standard & \textbf{16.7} & \textbf{31.3} & 58.6 & 15.4 & 21.8 & 28.8 \\
     & CoT & 13.3 & 28.7 & 58.8 & 16.2 & 22.4 & 27.9 \\
     & Plan-and-Solve & 10.0 & 30.1 & 56.8 & 18.4 & 20.4 & 27.1 \\
     & Hi-CoT (format-relaxed) & 6.7 & 30.1 & 63.0 & 18.4 & 23.9 & 28.4 \\
    \rowcolor{rowgray}
     & Hi-CoT & 13.3 & 27.7 & \textbf{63.4} & \textbf{19.1} & \textbf{26.1} & \textbf{29.9} \\
    \midrule
    \multirow{5}{*}{\textbf{\texttt{DeepSeek-R1-Distill-Qwen-7B}}} & Standard & 10.0 & \textbf{47.0} & 72.2 & 30.1 & 29.9 & 37.9 \\
     & CoT & 13.3 & 42.8 & 70.6 & 27.2 & 30.1 & 36.8 \\
     & Plan-and-Solve & \textbf{20.0} & 43.4 & 71.0 & 29.4 & 31.0 & 38.9 \\
     & Hi-CoT (format-relaxed) & 13.3 & \textbf{47.0} & 76.0 & 30.9 & 34.5 & 40.3 \\
    \rowcolor{rowgray}
     & Hi-CoT & 16.7 & 45.8 & \textbf{77.2} & \textbf{32.4} & \textbf{35.0} & \textbf{41.4} \\
    \midrule
    \multirow{5}{*}{\textbf{\texttt{DeepSeek-R1-Distill-Llama-8B}}} & Standard & 10.0 & \textbf{38.6} & 63.8 & 19.1 & 26.4 & 31.6 \\
     & CoT & 6.7 & 34.9 & 62.4 & 18.4 & 24.6 & 29.4 \\
     & Plan-and-Solve & 10.0 & 27.7 & 62.0 & 18.4 & 25.6 & 28.7 \\
     & Hi-CoT (format-relaxed) & 10.0 & 34.9 & 65.0 & 20.2 & 31.1 & 32.3 \\
    \rowcolor{rowgray}
     & Hi-CoT & \textbf{16.7} & 34.9 & \textbf{65.2} & \textbf{22.4} & \textbf{32.7} & \textbf{34.4} \\
    \midrule
    \multirow{5}{*}{\textbf{\texttt{DeepSeek-R1-Distill-Qwen-14B}}} & Standard & 13.3 & 50.6 & 73.2 & \textbf{36.0} & 34.2 & 41.5 \\
     & CoT & \textbf{16.7} & 38.6 & 72.0 & 26.8 & 34.1 & 37.6 \\
     & Plan-and-Solve & 10.0 & 41.0 & 71.8 & 33.8 & 33.5 & 38.0 \\
     & Hi-CoT (format-relaxed) & 13.3 & 51.4 & 78.2 & 35.3 & 35.7 & 42.8 \\
    \rowcolor{rowgray}
     & Hi-CoT & \textbf{16.7} & \textbf{51.8} & \textbf{78.4} & 35.7 & \textbf{39.3} & \textbf{44.4} \\
    \midrule
    \multirow{5}{*}{\textbf{\texttt{DeepSeek-R1-Distill-Qwen-32B}}} & Standard & 16.7 & 47.0 & 75.6 & 33.8 & 35.1 & 41.6 \\
     & CoT & 16.7 & 43.4 & 74.2 & 33.5 & 33.2 & 40.2 \\
     & Plan-and-Solve & \textbf{20.0} & 47.0 & 76.2 & 34.9 & 38.8 & 43.4 \\
     & Hi-CoT (format-relaxed) & 16.7 & \textbf{53.0} & 80.8 & 38.6 & 41.8 & 46.2 \\
    \rowcolor{rowgray}
     & Hi-CoT & 16.7 & 49.4 & \textbf{82.6} & \textbf{39.7} & \textbf{42.7} & \textbf{46.2} \\
    \bottomrule
  \end{tabular}%
  }
\end{table*}

\subsection{Experimental Setup}

\subsubsection{Baseline prompting methods.}
We compare our Hi-CoT with standard, CoT \citep{wei2022chain, kojima2022large} and Plan-and-Solve (PS) \citep{wang2023plan} prompting methods, as well as a Hi-CoT variant without the strict structure format named Hi-CoT (format-relaxed), across a range of LLMs on multiple mathematical reasoning benchmarks.
We focus on zero-shot prompting methods, as prior work suggests that CoT exemplars offer limited gains in reasoning for modern LLMs and primarily serve to enforce output format \citep{cheng2025revisiting}.
Prompt examples of the baseline methods can be found in the Appendix \ref{app:prompt}.

\subsubsection{Models}
We evaluate eight LLMs from the Qwen3 family \footnote{https://huggingface.co/collections/Qwen/qwen3}
 and five DeepSeek distilled models \footnote{https://huggingface.co/deepseek-ai/models}
: Qwen3-0.6B, 1.7B, 4B, 8B, and 32B; Qwen3-4B-Thinking and Qwen3-4B-Instruct; as well as DeepSeek-R1-Distill-Qwen-1.5B, 7B, 14B, and 32B, and DeepSeek-R1-Distill-Llama-8B. These models span a range of scales and specializations (e.g., instruction-tuned and reasoning-focused variants), allowing us to evaluate the robustness and generality of Hi-CoT across different model types.

\subsubsection{Tasks}
We evaluate our method on five widely used mathematical reasoning benchmarks: AIME24 \citep{li2024numinamath}, AMC \citep{li2024numinamath}, MATH500 \citep{hendrycks2021measuring}, Minerva Math \citep{lewkowycz2022solving}, and OlympiadBench \citep{he2024olympiadbench}. We focus on mathematical tasks because they require precise, multi-step reasoning with clearly verifiable outcomes, making them well-suited for evaluating both the accuracy and efficiency of structured reasoning methods. These benchmarks span a wide range of difficulty, from competition-level problems to advanced mathematical challenges, providing a comprehensive testbed for assessing reasoning performance.

\subsubsection{Evaluation Metrics}
We report Pass@1 accuracy as the primary metric, defined as the percentage of problems for which the model produces the correct final answer (formatted within \texttt{\textbackslash boxed$\left\{\right\}$}). For AIME24 \citep{li2024numinamath}, where answers are integers between 0 and 999, we follow standard evaluation practice and count only exact matches. Additionally, we measure the average length of reasoning tokens to assess the computational efficiency of Hi-CoT.

\subsection{Results}
In this part, we provide a comprehensive analysis of the reasoning performance of Hi-CoT compared to baseline methods, demonstrating its superior accuracy and computational efficiency.

\subsubsection{Performance Comparison}
Table \ref{tab:model_eval_results} presents the evaluation of Pass@1 accuracy (\%) on five diverse mathematical reasoning benchmarks. Across all 13 model configurations, Hi-CoT consistently achieves superior performance, with relative accuracy gains over both Standard and CoT prompting ranging from +3.5\% to +46.3\%. 
Compared with Plan-and-Solve, Hi-CoT also delivers stronger results on nearly all models and benchmarks, suggesting that explicitly separating and alternating instruction and execution provides a more effective form of structured reasoning.
These gains are consistent across model scales, from 0.6B to 32B parameters, indicating that the benefits of hierarchical reasoning are not tied to model size.
The effect is particularly pronounced for small models, i.e., Qwen3-4B-Thinking-2507, where Hi-CoT leads to substantial improvements. In this regime, the hierarchical structure appears to act as a scaffold, helping compensate for limited intrinsic reasoning capacity.
The improvements on DeepSeek distilled models are generally smaller than those observed on Qwen3 models. A possible explanation is that Qwen3 models exhibit stronger instruction-following behavior, making them more responsive to the structured prompting format imposed by Hi-CoT.

\begin{figure*}[ht]
  \centering
  \includegraphics[width=\textwidth]{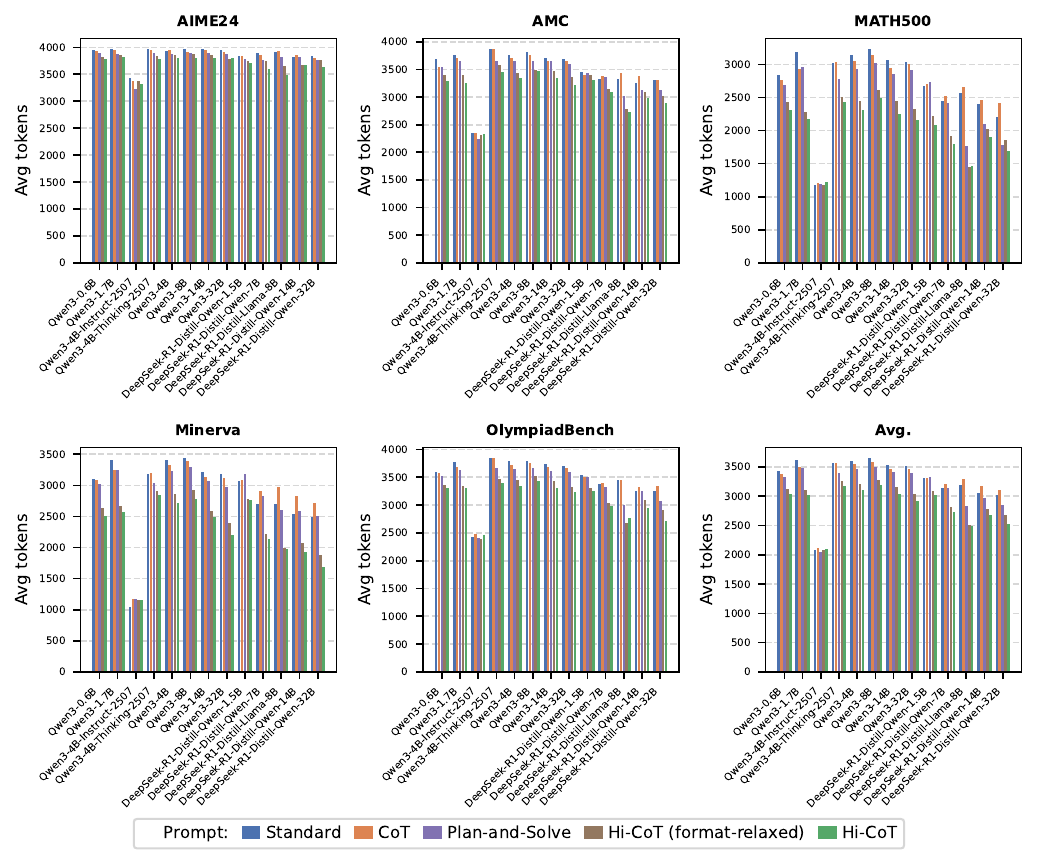}
  \caption{Average response length (tokens) for multiple prompting methods. 
  The rightmost panel shows the macro-average. 
  Lower value generally indicates higher efficiency.}
  \label{fig:avg_lens}
\end{figure*}

It can also be seen that Hi-CoT delivers the most pronounced gains on complex, multi-step reasoning tasks that demand structured decomposition. On the challenging AIME24 task, Hi-CoT lifts performance from near-zero accuracy (under Standard/CoT) to meaningful levels, with absolute gains of up to +20.0 points (e.g., Qwen3-14B: 3.3\% → 23.3\%). 
For MATH500 and OlympiadBench, Hi-CoT consistently yields absolute improvements across all models, as the explicit \texttt{<|instruction|>}/\texttt{<|execution|>} alternation guides models to systematically break down problems and avoid disorganized or redundant reasoning steps. 

We also note that Standard prompting already achieves relatively strong performance for some models. This may be due to their built-in “thinking” capabilities of modern LLM models, which can trigger implicit reasoning even without explicit guidance. As a result, CoT prompting often yields only marginal improvements or even worse performance over Standard prompting. 
Plan-and-Solve generally performs better than these two baselines, further supporting the importance of structured reasoning. However, Hi-CoT consistently surpasses all baselines, highlighting the advantage of explicitly enforcing a hierarchical alternating instruction–execution process.

\textbf{Ablation study.} Our Hi-CoT variant, namely Hi-CoT (format-relaxed), also shows improved performance over other baselines, while it still underperforms the fully constrained Hi-CoT. This suggests that introducing hierarchical reasoning guidance is beneficial, even without strict format enforcement. However, the performance gap indicates that explicitly enforcing the alternating \texttt{<|instruction|>} and \texttt{<|execution|>} structure plays a critical role in guiding the model’s behavior. In particular, the strict format reduces ambiguity in the reasoning process and helps maintain a clear separation between planning and execution, leading to more consistent and effective reasoning trajectories.

\begin{figure*}[ht]
  \centering
  \includegraphics[width=\textwidth]{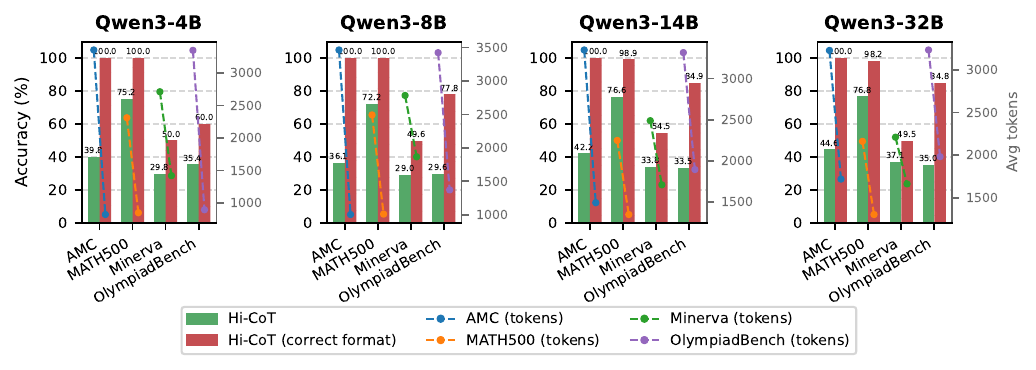}
  \caption{Comparative performance of Hi-CoT across aggregate and format-compliant outputs. Accuracy and average tokens are evaluated on four mathematical benchmarks for selected Qwen models, highlighting the performance gain when the hierarchical reasoning structure is strictly followed.}
  \label{fig:hier_format}
\end{figure*}


\subsubsection{Efficiency Evaluation}

We evaluate the efficiency of Hi-CoT by comparing its average response token length against the other baseline methods (Figure \ref{fig:avg_lens}). Across all benchmarks and model configurations, Hi-CoT consistently reduces response lengths, with relative reductions of 13.9 \% and up to 46.3\%. This trend is particularly evident on MATH500, where the structured hierarchical process eliminates redundant intermediate steps and disorganized tangents, reducing output by an average of 500–1200 tokens per response. Such reductions significantly lower both inference latency and computational overhead.
Notably, the token length remains relatively unreduced for the Qwen3-4B-Instruct-2507 model. We attribute this to the model’s lack of a native ``thinking'' mode, which results in a baseline reasoning length that is already minimal.

The efficiency gains scale consistently across model sizes, from 0.6B to 32B parameters, indicating that the token-reduction effect is agnostic to model scale and architecture.
Global averages confirm this pattern, demonstrating that hierarchical reasoning improves accuracy while simultaneously enhancing efficiency by prioritizing goal-oriented steps over unstructured exploration.




\subsubsection{Hierarchy Format Alignment}

We further evaluate the performance of responses that strictly adhere to the hierarchical format, defined as the accuracy achieved within the subset of properly formatted outputs. Format compliance is validated across three dimensions: 1) verification of alternating instruction and execution steps, 2) alignment of reasoning steps with the required structure, and 3) the presence of a final answer enclosed in \texttt{\textbackslash boxed$\left\{\right\}$}. 

As shown in Figure \ref{fig:hier_format}, the accuracy of conforming responses significantly exceeds the aggregate accuracy. Notably, on the AMC and MATH500 benchmarks, accuracy reaches 100\% when the model strictly follows the hierarchical framework. 
However, due to the high complexity of the AIME24 benchmark, some models fail to generate correctly formatted responses. Therefore, we report accuracy improvements only on the remaining four benchmarks.
Furthermore, as illustrated in the line charts of Figure \ref{fig:hier_format}, reasoning length is further reduced by up to 75.4\% when the model adheres to the structure. These findings underscore the critical role of explicit structural constraints in enabling LLMs to navigate complex reasoning trajectories efficiently.

A current bottleneck is the model's inconsistent adherence to the hierarchical format. Since our current experiments do not involve format-specific fine-tuning, the models occasionally revert to unstructured outputs, suggesting that supervised fine-tuning (SFT) or RL could further unlock the latent potential of the Hi-CoT paradigm.

\section{Conclusion}

In this work, we introduce Hierarchical Chain-of-Thought (Hi-CoT), a structured reasoning paradigm that transforms LLM reasoning from a flat sequence into a hierarchical, alternating instruct-and-execute process. Extensive experiments across 13 model configurations show that Hi-CoT consistently outperforms baseline prompting methods, achieving average accuracy gains of 6.2\% and up to 61.4\% on certain models and tasks, while reducing average token length by 13.9\%, effectively improving both reasoning quality and efficiency. 
Our analysis further reveals that when models strictly adhere to the hierarchical format, accuracy can reach 100\% on challenging benchmarks like AMC and MATH500, while further slashing reasoning traces by over 75\%. 
These results suggest that the primary bottleneck may not lie solely in the models' latent reasoning capacity, but could also be attributed to the relatively unstructured nature of current reasoning methods.
The consistent performance of Hi-CoT across various scales (0.6B to 32B) suggests its potential as a versatile and efficient framework for enhancing reasoning in LLMs.


Hi-CoT demonstrates strong performance and efficiency, but still has several important limitations. First, its effectiveness relies heavily on strict format adherence, and smaller models (e.g., Qwen3-0.6B and DeepSeek-R1-Distill-Qwen-1.5B) often fail to maintain the required structure, leading to only marginal improvements and indicating a minimum reasoning capability threshold. Second, its rigid hierarchical prompting design, while effective for structured tasks like mathematical reasoning, may limit flexibility and generalization to more open-ended domains. Finally, the current study is limited to mid-scale models (up to 32B) and prompt-level interventions, leaving uncertainty about its performance on larger frontier models and missing opportunities to further enhance results through post-training approaches such as SFT or RL tailored for hierarchical reasoning.





\bibliography{ref}
\bibliographystyle{icml2026}

\newpage
\appendix
\onecolumn
\section{Appendix}

\subsection{Baseline Prompting Methods}
\label{app:prompt}
This part shows the prompting examples of baseline methods.

\textbf{Standard prompting} provides the model with a task description and input, and expects it to produce the final answer directly, without any explicit reasoning trace. 

  \begin{stdbox}
  \small
  \textbf{Prompt:} Answer the question directly and put your final answer within \texttt{\textbackslash boxed$\left\{\right\}$}.\\[2pt]
  \textit{A store sells apples for \$3 each and oranges for \$5 each.
  If Alice buys 4 apples and 2 oranges, how much does she pay in total?}
  \tcblower
  \small
  \textbf{Output:} Alice pays \(\boxed{\$22}\).
  \end{stdbox}

\textbf{CoT prompting} \citep{wei2022chain, kojima2022large} extends standard prompting by explicitly encouraging the model to produce intermediate reasoning steps before arriving at a final answer. 

\begin{cotbox}
\small
\textbf{Prompt:} Please reason step by step, and put your final answer within \texttt{\textbackslash boxed$\left\{\right\}$}.\\[2pt]
\textit{A store sells apples for \$3 each and oranges for \$5 each.
If Alice buys 4 apples and 2 oranges, how much does she pay in total?}
 
\tcblower
\small
\textbf{Output:}\\
Alice buys 4 apples at \$3 each, so the cost for apples is $4 \times 3 = 12$ dollars. \\
She also buys 2 oranges at \$5 each, so the cost for oranges is $2 \times 5 = 10$ dollars. \\
The total cost is $12 + 10 = 22$ dollars. \\
Therefore, the final answer is \(\boxed{\$22}\).
\end{cotbox}

\textbf{Plan-and-Solve Prompting} \citep{wang2023plan} improves zero-shot CoT by first planning task decomposition and then solving subtasks step-by-step. Among its proposed six prompt templates, we select the one reporting the best performance.

\begin{psbox}
\small
\textbf{Prompt:} Let’s first understand the problem, extract relevant variables and their corresponding numerals, and make a complete plan. Then, let’s carry out the plan, calculate intermediate variables (pay attention to correct numerical calculation and common sense), solve the problem step by step, and finally put your final answer within \texttt{\textbackslash boxed$\left\{\right\}$}.\\[2pt]
\textit{A store sells apples for \$3 each and oranges for \$5 each.
If Alice buys 4 apples and 2 oranges, how much does she pay in total?}
\tcblower
\small
\textbf{Output:} First, let us identify the relevant quantities in the problem:

- Apples cost \$3 each

- Oranges cost \$5 each

- Alice buys 4 apples

- Alice buys 2 oranges

Plan:

1. Compute the total cost of the apples.

2. Compute the total cost of the oranges.

3. Add the two amounts to get the final total.

Now carry out the plan step by step.

Cost of apples:
$4 \times 3 = 12$

Cost of oranges:
$2 \times 5 = 10$

Total cost:
12 + 10 = 22

Therefore, Alice pays \(\boxed{22}\).
\end{psbox}

\textbf{Hi-CoT (format-relaxed)} removes the strict structure format from Hi-CoT.

\begin{hcotvbox}
\small
\textbf{Prompt:} You are a reasoning assistant that solves problems by alternating between \texttt{<|instruction|>} (planning what to do next) and \texttt{<|execution|>} (executing that plan).  
Each \texttt{<|instruction|>} describes the reasoning step or plan, and each \texttt{<|execution|>} performs the corresponding reasoning or computation. Finally, put your final answer within \texttt{\textbackslash boxed$\left\{\right\}$}.\\[2pt]
\textit{A store sells apples for \$3 each and oranges for \$5 each.
If Alice buys 4 apples and 2 oranges, how much does she pay in total?}
\tcblower
\small
\textbf{Output:} \texttt{<|instruction|>} Compute the total cost of the apples and the oranges separately.\\
\texttt{<|execution|>} The apples cost \(4 \times 3 = 12\), and the oranges cost \(2 \times 5 = 10\).\\
\texttt{<|instruction|>} Add the two partial costs to obtain the total amount Alice pays.\\
\texttt{<|execution|>} The total cost is \(12 + 10 = 22\). Therefore, Alice pays \(\boxed{\$22}\).
\end{hcotvbox}

\end{document}